\documentclass{article}
\usepackage{microtype}
\usepackage{graphicx}
\usepackage{subfigure}
\usepackage{booktabs} 
\usepackage{hyperref}

\usepackage[accepted]{icml2018}
\usepackage{latexsym}
\usepackage{amsfonts,amssymb}
\usepackage{amsmath}
\usepackage{graphicx}
\usepackage{algorithm}
\usepackage{algorithmic}
\usepackage{float}
\usepackage{array}
\usepackage{multirow}
\usepackage{url}
\usepackage{makecell}

\begin{document}

\twocolumn[
\icmltitle{Intelligence Graph}

\begin{icmlauthorlist}
\icmlauthor{Han Xiao}{thu}
\end{icmlauthorlist}

\icmlaffiliation{thu}{State Key Laboratory of Intelligent Technology and Systems, National Laboratory for Information Science and Technology, Department of Computer Science and Technology, Tsinghua  University, Beijing 100084, PR China}

\icmlcorrespondingauthor{Han Xiao}{Almighty.Xiao.Han@iCloud.com}

\icmlkeywords{Intelligence Graph}

\vskip 0.3in]

\printAffiliationsAndNotice{}  

\begin{abstract}
In fact, there exist three genres of intelligence architectures: logics (e.g. \textit{Random Forest, A$^*$ Searching}), neurons (e.g. \textit{CNN, LSTM}) and probabilities (e.g. \textit{Naive Bayes, HMM}), all of which are incompatible to each other. However, to construct powerful intelligence systems with various methods, we propose the intelligence graph (short as \textbf{\textit{iGraph}}), which is composed by both of neural and probabilistic graph, under the framework of forward-backward propagation. By the paradigm of iGraph, we design a recommendation model with semantic principle. First, the probabilistic distributions of categories are generated from the embedding representations of users/items, in the manner of neurons. Second, the probabilistic graph infers the distributions of features, in the manner of probabilities. Last, for the recommendation diversity, we perform an expectation computation then conduct a logic judgment, in the manner of logics. Experimentally, we beat the state-of-the-art baselines and verify our conclusions.
\end{abstract}

\section{Introduction}
In fact, there exist three genres of intelligence architectures: logics (e.g. \textit{Random Forest \cite{Zhou2017Deep}, A$^*$ Searching \cite{Munos2012The}}), neurons (e.g. \textit{CNN \cite{He2016Deep}, LSTM\cite{xiao2017hungarian}}) and probabilities (e.g. \textit{Naive Bayes \cite{Koller2009Probabilistic}, HMM \cite{Murphy2012Machine}}). The proposed neural graph \cite{xiao2017ndt} unifies the methodology of logics and neurons, which provides a more powerful form for intelligence systems. However, from the perspective of uncertainty, neural graph only characterizes the system in a deterministic way. In order to model intelligence in both deterministic and stochastic manner, we shall unify the neural and probabilistic graph in this paper as intelligence graph, or \textbf{\textit{iGraph}}. From the perspective of practice, the system designer could freely employ the models from both neural and probabilistic graph \cite{xiao2017ndt,Jordan2004Graphical} in one architecture at the same time.

\begin{figure}[H]
	\centering
	\includegraphics[width=0.8\linewidth]{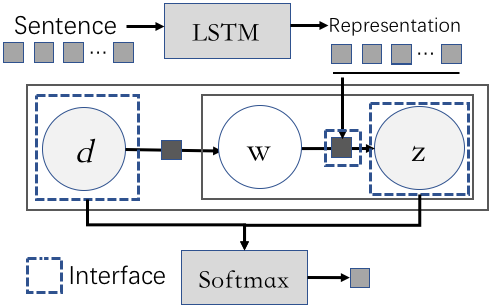}
	\caption{The illustration of the interface between neurons and probabilities for the task of text classification. To start, the sentence is parsed by LSTM to generate the context-based word representations, which are leveraged as $\mathcal{P}(z_k|w_i)$ in next probabilistic graph to calculate $\mathcal{P}(z_k|d_j)$. Last, a softmax layer is employed to classify the documents with the corresponding topic distributions.}
	\label{fig:show}
\end{figure}

Specifically, the neural and probabilistic parts corporate in the framework of forward-backward propagation, \cite{Rumelhart1988Learning, xiao2017hungarian}. In iGraph, we indicate the interface between neurons and probabilistic variables, where input interface links neurons to stochastic parts and output interface plays the otherwise corresponding role. As shown in Figure \ref{fig:show}, we leverage the topic model \cite{Murphy2012Machine} (notice the difference between pLSI) to classify documents, where the topic distribution of words $\mathcal{P}(z_k|w_i)$ stem from the hidden representations of LSTM, rather than the directly learned distribution vectors. Then, the soft-max layer is employed for the document classification with the corresponding topic distribution $\mathcal{P}(z_k|d_j)$. Mathematically, we could derive the system as below:
\begin{eqnarray}
	&& \mathcal{P}(z_k|w_i) \doteq softmax(\mathbf{Mh}_i),\;\; \mathbf{h}_i = \mathbf{LSTM}(\{w_i\}) \nonumber \\
	&& \mathcal{P}(z_k|d_j) = \sum_{w_i} \mathcal{P}(z_k|w_i) \mathcal{P}(w_i|d_j) \label{e5}\\
	&& y_j = softmax(\mathbf{W}\mathbf{f}_j),\;\;\mathbf{f}_j \doteq \mathcal{P}(z_k|d_j)
\end{eqnarray}
where $\mathbf{M, W}$ is the parameter of softmax and $\mathbf{LSTM}$ is the conventional network of LSTM.

\textit{\textbf{In essence, the gradients could be automatically derived, only if we could formulate the probabilistic part}}. Simply, in the forward pass, we treat the input/output interface as input/output probabilistic distribution, while \textit{\textbf{the sum-product algorithm formulates this process for calculation}}, \cite{Kschischang2001Factor}. For the example of Figure \ref{fig:show}, the input/output distribution is $\mathcal{P}(z_k|w_i)$/$\mathcal{P}(z_k|d_j)$. To start, we work out the joint probabilistic distribution as $\mathcal{P}(z_k, w_i, d_j) = \mathcal{P}(z_k|w_i)\mathcal{P}(w_i|d_j)$, and then the formulation of sum-product algorithm is presented in Equation (\ref{e5}). In the backward pass, the gradient propagates though Equation (\ref{e5}) to LSTM, in a conventional manner. \textit{\textbf{Notably, given the specific iGraph, all of the deductions could be performed automatically.}}

\begin{figure}[t]
	\centering
	\includegraphics[width=\linewidth]{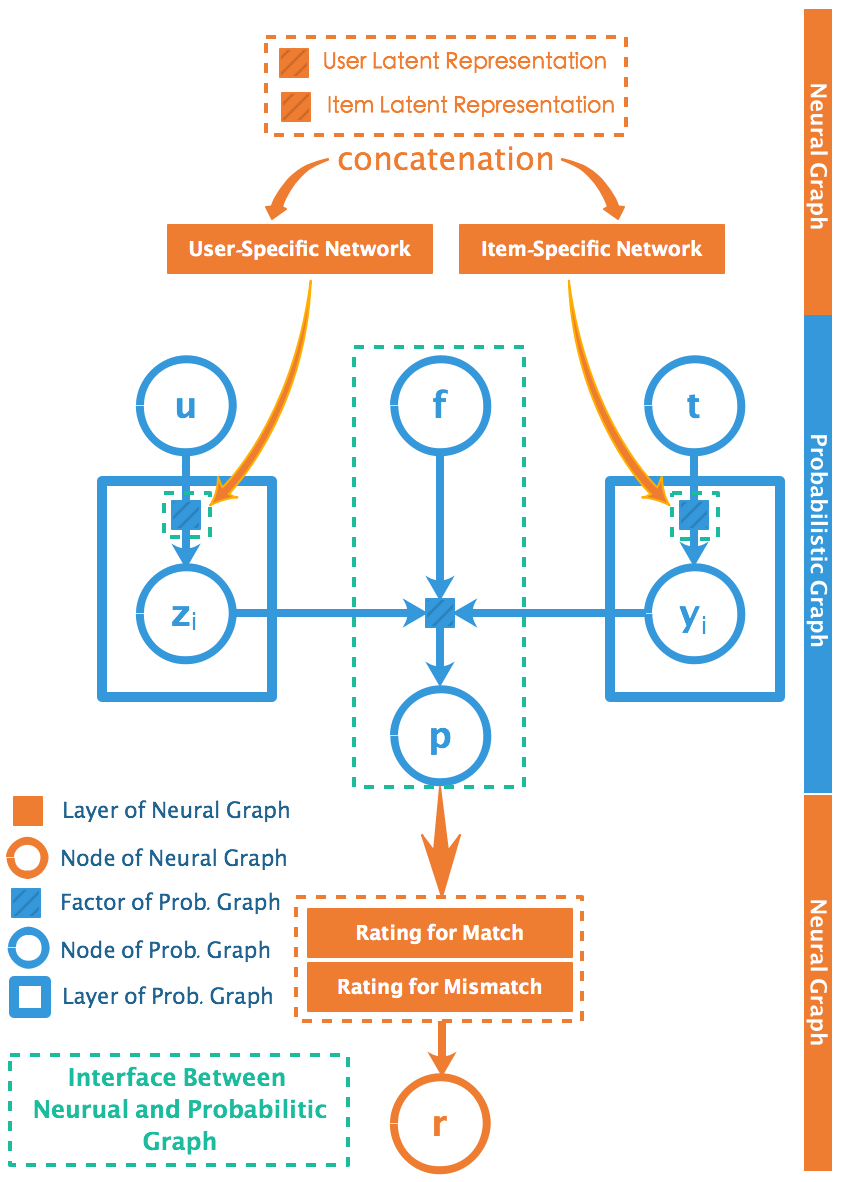}
	\caption{The iGraph of Our Model. First, the probabilistic distributions of categories are generated from the embedding representations of users/items, in the manner of neurons. Second, the probabilistic graph infers the distributions of features, in the manner of probabilities. Last, for the diversity of recommendation, we perform an expectation computation then conduct a logic judgment, in the manner of logics.}
	\label{fig:igraph}
\end{figure}

With this principle of iGraph, we tackle the task of recommendation in Figure \ref{fig:igraph}, as an extension of SAR \cite{Xiao2017SAR}. SAR is a probabilistic model, which applies semantic principle \cite{xiao2016knowledge}, to represent users/items semantically, and then treats the recommendation as a process of semantic matching. There exist two disadvantages for SAR, listed as \textbf{(1.)} The characterization of category distribution that a distribution vector is too simplified to achieve better performance. \textbf{(2.)} There is no mechanism for considering recommendation diversity. 

Regarding the first disadvantage, we employ a neural network to generate the category distributions from user and item embedding representations. In this way, the category distributions are modeled precisely. Regarding the second disadvantage, we respectively process the situations where the predicted rating is high or medium. Actually, a war-movie fan would not expect all the recommended films are war-related, because \textit{\textbf{some high-quality movies would not make a perfect semantic match, but still could be highly rated}}. In the final stage of our iGraph, for the diversity of recommendation, we check out the case of high-quality movies and make an appropriate recommendation based on the quality and popularity.

We conduct the experiments of rating prediction on  Movielens dataset to verify our model. Experimental results demonstrate the effectiveness and efficiency of our model, because our model beats all the state-of-the-art baselines. We also vary the hyper-parameters to testify our model, and conclude that our model is robust for hyper-parameter settings.  

\textbf{Contributions.} We list two contributions: \textbf{(1.)} \textit{\textbf{Intelligence graph (iGraph) could represent all the combinations and iterations of almost every intelligence method, which incurs a complete representation theory of intelligence.}} \textbf{(2.)} We design a novel graph for recommendation based on SAR and tackle two issues that the oversimplified category distribution and recommendation diversity. Besides, we achieve the state-of-the-art performance in the task of rating prediction.

\section{Related Work}
We have surveyed the relevant papers and roughly classified the existing recommendation methods into five categories: \textit{matrix factorization, neighborhood based method, regression based method, social information integration method and semantic analysis}. Notably, most methods are based on rating matrix completion, which is a conventional setting for recommendation, \cite{Xiao2017SAR}.

\textbf{Matrix Factorization} is a classic recommendation methodology. First, this paradigm  factorizes the rating matrix $M \approx UV$ to get the user/item-specific latent matrices $U$,$V$. Then, the method multiplies $UV \approx \hat{M}$ to estimate the unobservable ratings, where $\hat{M}$ is the complete rating matrix. Because this branch hypothesizes different assumptions on latent matrices $U$ and $V$, there also exist four primary subcategories according to the applied hypotheses. \textit{(1.) Basic matrix factorization} generally constrains latent factors as positive/non-negative, such as  NMF \cite{Wang2013Nonnegative}, SVP \cite{Meka2009Guaranteed}, MMMF \cite{Rennie2005Fast}, PMF \cite{Mnih2012Probabilistic}.  \textit{(2.) Matrix factorization jointed with neighborhood-based modeling}, such as CISMF \cite{Guo2015A}. \textit{(3.) Matrix factorization under various rank assumptions} explores the effects of representation and generalization ability for the matrix rank to boost the performance, such as LLORMA \cite{Lee2016LLORMA}\cite{Ko2015Multi}, R1MP \cite{Wang2014Rank}, SoftImpute \cite{Rahul2010Spectral}, \cite{Ganti2015Matrix}, \cite{Zhang2013Localized}, \cite{Kir2012The}. \textit{(4.) Matrix factorization with discrete outputs} treats each item of the rating matrix as discrete value to avoid noise and obtains more interpretations, such as ODMC \cite{Huo2016Optimal}.

\textbf{Neighborhood Based Method} is one of the most seminal approaches, assuming that the similar items/users trigger similar rating preference. There exist three main variants including item-based, user-based and global similarity, surveyed in \cite{Guo2015A} and \cite{Ricci2011Recommender}.

\textbf{Regression Based Method} formulates recommendation or matrix completion as a regression problem, such as graph regression method GRALS \cite{Cai2011Graph},  blind regression \cite{Song2016Blind}, Riemannian manifold based regression \cite{Vandereycken2013Low}, and others \cite{Davenport20141}.

\textbf{Social Information Integration} applies social information to strengthen the recommendation such as relationship between users, personalized profiles or movies' attributes. There list some latest researches: SR \cite{Ma2013An}, PRMF \cite{Liu2016Learning}, geo-specific personalization \cite{Liu2014Personalized}, social network based methods \cite{Deng2014Social} and other social context integration methods \cite{Wang2014Recommendation}.

\textbf{Semantic Analysis} takes the advantage of semantic principle or multi-view clustering methodology for recommendation \cite{Xiao2017SAR}. Semantic principle conjectures that clusters and semantics are equivalently corresponded, \cite{xiao2016knowledge}. Simply, SAR \cite{Xiao2017SAR} clusters the users/items in different views, where each view corresponds to a specific semantic style. Then, summarizing the cluster information in each semantic view, we obtain the user/item-specific semantic representation. Last, SAR performs a process of semantic matching to discriminate the highly rated items for users (better match triggers higher rating). This method is based on probabilistic graph, thus lacking of non-linear function and complex neural network structures leads to unsatisfactory performance. Also, there exist the recommendation diversity issue as previously discussed.

\section{Methodology}
Our iGraph is illustrated in Figure \ref{fig:igraph}. In this paper, we design a recommendation model with semantic principle. First, the probabilistic distributions of categories are generated from the embedding representations of users/items, in the manner of neurons. Second, the probabilistic graph infers the distributions of features, in the manner of probabilities. Last, for the recommendation diversity, we perform an expectation computation then conduct a logic judgment, in the manner of logics. 

\subsection{Architecture}
In the sequel, we discuss our graph in corresponding five components: embedding layer, user/item-specific network, semantic component, rating generation and logic judgment.

\textbf{Embedding Layer.} It is necessary to represent users/items in a latent manner, because the input of recommendation is oversimple. Actually, most recommendation methods take the idea of embedding. For the example of PMF, the row/column of factor matrices represent the user/item in a manner of probabilistic distribution. Specifically, regarding SAR, the category distribution of $\mathcal{P}(z_k|u)$/$\mathcal{P}(z_k|t)$ corresponds to the functionality of embedding representation. In this paper, we take the $k$-dimensional real vector as our embedding, and then we concatenate the corresponding user and item embedding vector as the embedding of this entry $M_{u,t}$.

\textbf{User/Item-Specific Network.} SAR simply supposes the user/item category distribution is only related to the corresponding user/item. However, we conjecture user category distribution should vary slightly with the different items, and similar case for item. By the flexibility provided by iGraph, we employ two specific networks to transform the embedding of rating matrix entry $M_{u,i}$ into the corresponding category distributions, which is the input interface of probabilistic part. Notably, we customize the network as multiple layer perceptions in the hyper-parameter setting.

\begin{figure*}
	\begin{eqnarray}
	&& \mathcal{P}(\hat{p}, z_n, y_n, f_n|u,t)  = \mathcal{P}(z_n|u, f_n) \mathcal{P}(y_n|t, f_n) \mathcal{P}(\hat{p}|z_n, y_n, f_n, u, t) \label{e1}
	\end{eqnarray}
	\begin{small}
		\fontsize{8pt}{0}
		\begin{eqnarray}
		& \mathcal{P}(\hat{p}|f, u,t) & =
		\overbrace{\sum_{n=1}^{|F|} \underline{\mathcal{P}(f_n|u,t)}
			\overbrace{\sum_{i,j=1}^{|C|}  \underline{\mathcal{P}(z_n=i|u, f_n) \mathcal{P}(y_n=j|t, f_n)} \mathcal{P}(\hat{p}|z_n=i, y_n=j, f_n, u, t) 
			}^{Second-Level:~Category~Mixture}
		}^{First-Level:~Feature~Mixture} \label{e2} \\
		& & =
		\sum_{n=1}^{|F|} \mathcal{P}(f_n|u,t)
		\sum_{i,j=1}^{|C|}  \mathcal{P}(z_n=i | u, f_n) \mathcal{P}(y_n=j|t, f_n) \tau_{\hat{p}|z_n, y_n, f_n} e^{-\frac{| \mathcal{P}(z_n =i| u, f_n) - \mathcal{P}(y_n=j | t, f_n)|}{\sigma}} \nonumber \\ \label{e3}
		\end{eqnarray}
	\end{small}	
\end{figure*}

\textbf{Semantic Component.} This component is a direct copy of SAR, which is a two-level hierarchical generation process. Simply, the model generates different features in the first-level process, while the user/item generates the categories for each feature in the second-level process, correspondingly. Finally, the categories in each feature generate the preference $p \in [1...|R|]$, where $|R|$ is the range of rating. There introduce three factors: $\mathcal{P}(z_i|u)$, $\mathcal{P}(y_i|t)$ and $\mathcal{P}(p|z_i, y_i)$, two of which are described in the previous paragraph. Regarding the remaining distribution $\mathcal{P}(p|z_i. y_i)$, we should consider the the effect of semantic matching as:
\begin{eqnarray}
\mathcal{P}(\hat{p}|z_n, y_n, f_n, u, t) =
\tau_{\hat{p}|z_n, y_n, f_n} e^{-\frac{| \mathcal{P}(z_n | u, f_n) - \mathcal{P}(y_n | t, f_n)|}{\sigma}} \label{e0}
\end{eqnarray}
where $\tau_{\hat{p}|z_n, y_n, f_n}$ is tabular model parameters which can be
tuned in the learning process and $\sigma$ is the hyper-parameter. 

Though the deduction of probabilistic part is automatically performed by sum-product rule, we still present the corresponding probabilistic form for clarity, where two-level mixture indicates the process of multi-view clustering methodology, according to semantic principle.

\textbf{Rating Generation.} Regarding the task of prediction, according to SAR,  rating is estimated as the expectation of soft-max distribution, as formulated in (\ref{e4}).
\begin{eqnarray}
r_{|u,t} \doteq \mathbb{E}_{r|u,t}(r) =  \frac{\sum_{p=1}^{|R|} p \times e^{\hat{p} \omega_u \omega_t}}{\sum_{p=1}^{|R|} e^{\hat{p}_i \omega_u \omega_t}} \label{e4}
\end{eqnarray}
where $p$ corresponds to the rating range, and $\hat{p}$ is the output interface as $\mathcal{P}(p|f,u,t)$, which has $|R|$ entries.

\begin{algorithm}[H]
	\renewcommand{\algorithmicrequire}{\textbf{Input:}}
	\renewcommand{\algorithmicensure}{\textbf{Output:}}
	\caption{Diversity Recommendation.}
	\label{alg}
	\begin{algorithmic}[1]
		\REQUIRE Predicted ratings $r_{|u,t}$, hyper-parameter $[r_a, r_b]$, feature distribution $\mathcal{P}(\hat{p}|f,u,t)$ and another item embedding $\mathbf{t_{diversity}}$.
		\ENSURE Final predicted rating $\hat{r}_{|u,t}$.
		\STATE $\hat{r}_{|u,t} = r_{|u,t}$
		\IF {$r_{|u,t} \in [r_a, r_b]$}
		\STATE $\hat{r}_{|u,t} = r_{|u,t} + NN(\mathbf{t_{diversity}}, \mathcal{P}(\hat{p}|f,u,t))$
		\ENDIF
		\STATE \textbf{return} $\hat{r}_{|u,t}$
	\end{algorithmic}
\end{algorithm}

\textbf{Logic Judgment.} Argued in the ``Introduction'', we should consider the high-quality movies for recommendation diversity. It is necessary to judge the match degree, because we only perform diversity recommendation for a special range of predicted rating, such as $[r_a, r_b]$. In fact, the highly rated items need no extra processing, thus we limit the upper bound as $r_a$, while extremely semantically unmatched items should not be considered as recommended, even if it is popular or high-quality, hence we limit the lower bound as $r_b$. Notice that $r_a, r_b$ are two distinguishable hyper-parameters.

Generally, we complement the rating in the range $[r_a, r_b]$ with a neural network as $\hat{r}_{u,t} = r_{|u,t} + r_{diversity}$, where $r_{diversity}$ is the output of the corresponding network and the input is another item embedding $\mathbf{t_{diversity}}$ and feature distribution $\mathcal{P}(\hat{p}|f,u,t)$. To summarize, we present this process in Algorithm \ref{alg}. Notice that the network is customized as a multiple layer perception in the hyper-parameter setting.

\subsection{From the Perspective of Interpretability}
Commonly agreed by our community, neural networks or neural parts are black-box. The critical point of neural style is strong data-fitting ability, while the flaw is lacking of interpretability.

However, a better interpretability takes at least three advantages. First, a good intuition inspires a better architecture. Well-defined interpretability may even promote the performance in the breakthrough level. Second, many areas need the cooperation between machines and humans, where interpretability is a necessary option. Last, interpretability could joint with handy work such as rules, which opens an industrial way for intelligence system. Thus, logic and probabilistic part, which could provide strong interpretability, are critical in intelligence graph. 

On the other side, the methods from traditional algorithms and probabilistic graphs perform less satisfactory than neural networks, because of a weaker data-fitting ability. Data-fitting ability, which brings the research trend of deep learning, is also critical for intelligence systems. In summary, we shall jointly consider the data-fitting ability and interpretability in the intelligence graph.

In my opinion, interpretable (i.e. \textit{probabilistic and logic}) parts should dominate the overview framework of iGraph, and neural parts are responsible for feature extraction and links  between the interpretable parts. In this way, the entire architecture is interpretable with strong data-fitting ability.

\section{Experiment}
This section has not been ready in this version.

\section{Conclusion}
In this paper, we propose a complete representation theory of intelligence, as intelligence graph \textbf{\textit{(iGraph)}}. Based on this novel paradigm, we design a graph for recommendation to tackle two issues, that the oversimple category distribution and diversity recommendation. Experimental results demonstrate the effectiveness and efficiency of our model.

\newpage

\bibliography{Ref}

\begin{thebibliography}{35}
\providecommand{\natexlab}[1]{#1}
\providecommand{\url}[1]{\texttt{#1}}
\expandafter\ifx\csname urlstyle\endcsname\relax
  \providecommand{\doi}[1]{doi: #1}\else
  \providecommand{\doi}{doi: \begingroup \urlstyle{rm}\Url}\fi

\bibitem[Cai et~al.(2011)Cai, He, Han, and Huang]{Cai2011Graph}
Cai, Deng, He, Xiaofei, Han, Jiawei, and Huang, Thomas~S.
\newblock Graph regularized nonnegative matrix factorization for data
  representation.
\newblock \emph{Pattern Analysis and Machine Intelligence IEEE Transactions
  on}, 33\penalty0 (8):\penalty0 1548--1560, 2011.

\bibitem[Davenport et~al.(2014)Davenport, Plan, Berg, and
  Wootters]{Davenport20141}
Davenport, Mark~A., Plan, Yaniv, Berg, Ewout Van~Den, and Wootters, Mary.
\newblock 1-bit matrix completion.
\newblock \emph{Statistics}, 3\penalty0 (3), 2014.

\bibitem[Deng et~al.(2014)Deng, Huang, and Xu]{Deng2014Social}
Deng, Shuiguang, Huang, Longtao, and Xu, Guandong.
\newblock Social network-based service recommendation with trust enhancement.
\newblock \emph{Expert Systems with Applications}, 41\penalty0 (18):\penalty0
  8075--8084, 2014.

\bibitem[Ganti et~al.(2015)Ganti, Balzano, and Willett]{Ganti2015Matrix}
Ganti, Ravi, Balzano, Laura, and Willett, Rebecca.
\newblock Matrix completion under monotonic single index models.
\newblock 2015.

\bibitem[Guo et~al.(2015)Guo, Sun, and Meng]{Guo2015A}
Guo, Meng~Jiao, Sun, Jin~Guang, and Meng, Xiang~Fu.
\newblock A neighborhood-based matrix factorization technique for
  recommendation.
\newblock \emph{Annals of Data Science}, 2\penalty0 (3):\penalty0 1--16, 2015.

\bibitem[He et~al.(2016)He, Zhang, Ren, and Sun]{He2016Deep}
He, Kaiming, Zhang, Xiangyu, Ren, Shaoqing, and Sun, Jian.
\newblock Deep residual learning for image recognition.
\newblock In \emph{Computer Vision and Pattern Recognition}, pp.\  770--778,
  2016.

\bibitem[Huo et~al.(2016)Huo, Liu, and Huang]{Huo2016Optimal}
Huo, Zhouyuan, Liu, Ji, and Huang, Heng.
\newblock Optimal discrete matrix completion.
\newblock 2016.

\bibitem[Jordan(2004)]{Jordan2004Graphical}
Jordan, Michael~I.
\newblock Graphical models.
\newblock \emph{Statistical Science}, 19\penalty0 (1):\penalty0 140--155, 2004.

\bibitem[KirÃ¡ly et~al.(2012)KirÃ¡ly, Theran, and Tomioka]{Kir2012The}
KirÃ¡ly, Franz~J., Theran, Louis, and Tomioka, Ryota.
\newblock The algebraic combinatorial approach for low-rank matrix completion.
\newblock \emph{Journal of Machine Learning Research}, 62\penalty0
  (2):\penalty0 299--321, 2012.

\bibitem[Ko et~al.(2015)Ko, Son, and Ko]{Ko2015Multi}
Ko, Han~Gyu, Son, Joo~Sik, and Ko, In~Young.
\newblock Multi-aspect collaborative filtering based on linked data for
  personalized recommendation.
\newblock In \emph{The International Conference on World Wide Web}, pp.\
  49--50, 2015.

\bibitem[Koller \& Friedman(2009)Koller and Friedman]{Koller2009Probabilistic}
Koller, Daphne and Friedman, Nir.
\newblock \emph{Probabilistic Graphical Models: Principles and Techniques -
  Adaptive Computation and Machine Learning}.
\newblock MIT Press, 2009.

\bibitem[Kschischang et~al.(2001)Kschischang, Frey, and
  Loeliger]{Kschischang2001Factor}
Kschischang, Frank~R, Frey, Brendan~J, and Loeliger, Hans~Andrea.
\newblock Factor graphs and the sum-product algorithm.
\newblock \emph{IEEE Transactions on Information Theory}, 47\penalty0
  (2):\penalty0 498--519, 2001.

\bibitem[Lee et~al.(2016)Lee, Kim, Lebanon, Singer, and Bengio]{Lee2016LLORMA}
Lee, J., Kim, S., Lebanon, G., Singer, Y., and Bengio, S.
\newblock Llorma: Local low-rank matrix approximation.
\newblock 2016.

\bibitem[Liu et~al.(2014)Liu, Li, Tang, and Jiang]{Liu2014Personalized}
Liu, Jing, Li, Zechao, Tang, Jinhui, and Jiang, Yu.
\newblock Personalized geo-specific tag recommendation for photos on social
  websites.
\newblock \emph{IEEE Transactions on Multimedia}, 16\penalty0 (3):\penalty0
  588--600, 2014.

\bibitem[Liu et~al.(2016)Liu, Zhao, Liu, Wu, and Li]{Liu2016Learning}
Liu, Yong, Zhao, Peilin, Liu, Xin, Wu, Min, and Li, Xiao~Li.
\newblock Learning optimal social dependency for recommendation.
\newblock 2016.

\bibitem[Ma(2013)]{Ma2013An}
Ma, Hao.
\newblock An experimental study on implicit social recommendation.
\newblock In \emph{International ACM SIGIR Conference on Research and
  Development in Information Retrieval}, pp.\  73--82, 2013.

\bibitem[Meka et~al.(2009)Meka, Jain, and Dhillon]{Meka2009Guaranteed}
Meka, Raghu, Jain, Prateek, and Dhillon, Inderjit~S.
\newblock Guaranteed rank minimization via singular value projection.
\newblock \emph{Nips}, pp.\  937--945, 2009.

\bibitem[Mnih \& Salakhutdinov(2012)Mnih and
  Salakhutdinov]{Mnih2012Probabilistic}
Mnih, A. and Salakhutdinov, R.
\newblock Probabilistic matrix factorization.
\newblock In \emph{International Conference on Machine Learning}, pp.\
  880--887, 2012.

\bibitem[Munos(2012)]{Munos2012The}
Munos, R.
\newblock The optimistic principle applied to games, optimization and planning:
  Towards foundations of monte-carlo tree search.
\newblock \emph{Foundations and Trends in Machine Learning}, 7\penalty0
  (1):\penalty0 1--130, 2012.

\bibitem[Murphy(2012)]{Murphy2012Machine}
Murphy, Kevin~P.
\newblock \emph{Machine Learning: A Probabilistic Perspective}.
\newblock MIT Press, 2012.

\bibitem[Rahul~Mazumder(2010)]{Rahul2010Spectral}
Rahul~Mazumder, Trevor~Hastie, Robert~Tibshirani.
\newblock Spectral regularization algorithms for learning large incomplete
  matrices.
\newblock \emph{Journal of Machine Learning Research}, 11\penalty0
  (11):\penalty0 2287--2322, 2010.

\bibitem[Rennie \& Srebro(2005)Rennie and Srebro]{Rennie2005Fast}
Rennie, Jasson D.~M and Srebro, Nathan.
\newblock Fast maximum margin matrix factorization for collaborative
  prediction.
\newblock In \emph{International Conference}, pp.\  713--719, 2005.

\bibitem[Ricci et~al.(2011)Ricci, Rokach, Shapira, and
  Kantor]{Ricci2011Recommender}
Ricci, Francesco, Rokach, Lior, Shapira, Bracha, and Kantor, Paul~B.
\newblock \emph{Recommender systems handbook /}.
\newblock Springer,, 2011.

\bibitem[Rumelhart et~al.(1988)Rumelhart, Hinton, and
  Williams]{Rumelhart1988Learning}
Rumelhart, D.~E., Hinton, G.~E., and Williams, R.~J.
\newblock \emph{Learning internal representations by error propagation}.
\newblock MIT Press, 1988.

\bibitem[Song(2016)]{Song2016Blind}
Song, Dogyoon.
\newblock Blind regression : nonparametric regression for latent variable
  models via collaborative filtering.
\newblock 2016.

\bibitem[Vandereycken(2013)]{Vandereycken2013Low}
Vandereycken, Bart.
\newblock Low-rank matrix completion by riemannian optimization.
\newblock \emph{Siam Journal on Optimization}, 23\penalty0 (23):\penalty0
  10.1137/110845768, 2013.

\bibitem[Wang et~al.(2014{\natexlab{a}})Wang, Jiang, Zhu, Yang, and
  Cui]{Wang2014Recommendation}
Wang, Fei, Jiang, Meng, Zhu, Wenwu, Yang, Shiqiang, and Cui, Peng.
\newblock Recommendation with social contextual information.
\newblock \emph{IEEE Transactions on Knowledge and Data Engineering},
  26\penalty0 (11):\penalty0 2789--2802, 2014{\natexlab{a}}.

\bibitem[Wang \& Zhang(2013)Wang and Zhang]{Wang2013Nonnegative}
Wang, Yu~Xiong and Zhang, Yu~Jin.
\newblock Nonnegative matrix factorization: A comprehensive review.
\newblock \emph{IEEE Transactions on Knowledge and Data Engineering},
  25\penalty0 (6):\penalty0 1336--1353, 2013.

\bibitem[Wang et~al.(2014{\natexlab{b}})Wang, Lai, Lu, Fan, Davulcu, and
  Ye]{Wang2014Rank}
Wang, Z., Lai, M.~J., Lu, Z., Fan, W., Davulcu, H., and Ye, J.
\newblock Rank-one matrix pursuit for matrix completion.
\newblock pp.\  91--99, 2014{\natexlab{b}}.

\bibitem[Xiao(2016)]{xiao2016knowledge}
Xiao, Han.
\newblock {KSR}: A semantic representation of knowledge graph within a novel
  unsupervised paradigm.
\newblock \emph{arXiv preprint arXiv:1608.07685}, 2016.

\bibitem[Xiao(2017{\natexlab{a}})]{xiao2017hungarian}
Xiao, Han.
\newblock Hungarian layer: Logics empowered neural architecture.
\newblock \emph{arXiv preprint arXiv:1712.02555}, 2017{\natexlab{a}}.

\bibitem[Xiao(2017{\natexlab{b}})]{xiao2017ndt}
Xiao, Han.
\newblock {NDT}: Neual decision tree towards fully functioned neural graph.
\newblock \emph{arXiv preprint arXiv:1712.05934}, 2017{\natexlab{b}}.

\bibitem[Xiao \& Meng(2017)Xiao and Meng]{Xiao2017SAR}
Xiao, Han and Meng, Lian.
\newblock Sar: Semantic analysis for recommendation.
\newblock 2017.

\bibitem[Zhang et~al.(2013)Zhang, Zhang, Liu, Ma, and Feng]{Zhang2013Localized}
Zhang, Yongfeng, Zhang, Min, Liu, Yiqun, Ma, Shaoping, and Feng, Shi.
\newblock Localized matrix factorization for recommendation based on matrix
  block diagonal forms.
\newblock 2013.

\bibitem[Zhou \& Feng(2017)Zhou and Feng]{Zhou2017Deep}
Zhou, Zhi~Hua and Feng, Ji.
\newblock Deep forest: Towards an alternative to deep neural networks.
\newblock 2017.

\end{thebibliography}
\bibliographystyle{icml2018}

\end{document}